\documentclass[letterpaper, 10 pt, conference]{ieeeconf}  % Comment this line out if you need a4paper

\IEEEoverridecommandlockouts                              % This command is only needed if 
\usepackage{graphics} % for pdf, bitmapped graphics files
\usepackage{epsfig} % for postscript graphics files
\usepackage{cite,color}
\usepackage{multirow}
\usepackage{pifont}
\usepackage{amsmath} % assumes amsmath package installed
\usepackage{amssymb,bm}  % assumes amsmath package installed
\usepackage{hyperref}
\hypersetup{hypertex=true}
\usepackage{booktabs}
\usepackage{etoolbox}
\makeatletter
\patchcmd{\@makecaption}
  {\scshape}
  {}
  {}
  {}
\patchcmd{\@makecaption}
  {\\}
  {.\ }
  {}
  {}
\makeatother

\usepackage{caption}
\title{\LARGE \bf
Time-optimal Flight in Cluttered Environments\\ via Safe Reinforcement Learning
}

\author{Wei Xiao, Zhaohan Feng, Ziyu Zhou, Jian Sun, Gang Wang, and Jie Chen % <-this % stops a space
    \thanks{The work was partially supported by the National Natural Science Foundation of China under Grants 62173034, U23B2059, 61925303, 62088101, and U22B2058.}% <-this % stops a space
    \thanks{Wei Xiao, Zhaohan Feng, Ziyu Zhou, Jian Sun, Gang Wang, and Jie Chen are with the National Key Lab of Autonomous Intelligent Unmanned Systems, Beijing Institute of Technology, Beijing 100081, China, and the Beijing Instituteof Technology Chongqing Innovation Center, Chongqing 401120, China.
    (email: xiaowei@bit.edu.cn, zhfeng@bit.edu.cn, ziyuzhou@bit.edu.cn, gangwang@bit.edu.cn, sunjian@bit.edu.cn). Jie Chen is with the National Key Lab of Autonomous Intelligent Unmanned Systems, Tongji University, Shanghai 201804, China (e-mail: chenjie@bit.edu.cn). %
    }
}

\begin{document}

\maketitle
\thispagestyle{empty}
\pagestyle{empty}

%%%%%%%%%%%%%%%%%%%%%%%%%%%%%%%%%%%%%%%%%%%%%%%%%%%%%%%%%%%%%%%%%%%%%%%%%%%%%%%%
\begin{abstract}
This paper addresses the problem of guiding a quadrotor through a predefined sequence of waypoints in cluttered environments, aiming to minimize the flight time while avoiding collisions.
Previous approaches either suffer from prolonged computational time caused by solving complex non-convex optimization problems or are limited by the inherent smoothness of polynomial trajectory representations, thereby restricting the flexibility of movement.
In this work, we present a safe reinforcement learning approach for autonomous drone racing with time-optimal flight in cluttered environments.
The reinforcement learning policy, trained using  safety and terminal rewards specifically designed to enforce near time-optimal and collision-free flight, outperforms current state-of-the-art algorithms.
Additionally, experimental results demonstrate the efficacy of the proposed approach in achieving both minimum flight time and obstacle avoidance objectives in complex environments, with a commendable $66.7\%$ success rate in unseen, challenging settings.
%This study offers a promising advancement in quadrotor navigation, with practical implications for real-world applications.

\end{abstract}

%%%%%%%%%%%%%%%%%%%%%%%%%%%%%%%%%%%%%%%%%%%%%%%%%%%%%%%%%%%%%%%%%%%%%%%%%%%%%%%%
\section{Introduction}

In recent years, drones have gained increasing popularity across various domains due to their exceptional performance, sparking a research fervor in autonomous drone racing. 
For instance, the AlphaPilot Challenge \cite{guerra2019flightgoggles}, \cite{foehn2022alphapilot}, the autonomous drone racing at IROS and NeurIPS \cite{auto_droneracing1,NeurIPSracing}. %auto_droneracing2,Beautyandthebeast, 
Autonomous drone racing is an excellent platform for challenging perception, trajectory planning and control technologies. 
The rules stipulate that the drones must systematically navigate through a sequence of gates, ensuring collision-free traversal while preserving the aggressive capabilities of the drones throughout the course. 

\begin{figure}[thpb]
        \centering
        %\framebox{\parbox{3in}{We suggest that you use a text box to insert a graphic (which is ideally a 300 dpi TIFF or EPS file, with all fonts embedded) because, in an document, this method is somewhat more stable than directly inserting a picture.}}
        \includegraphics[scale=0.25]{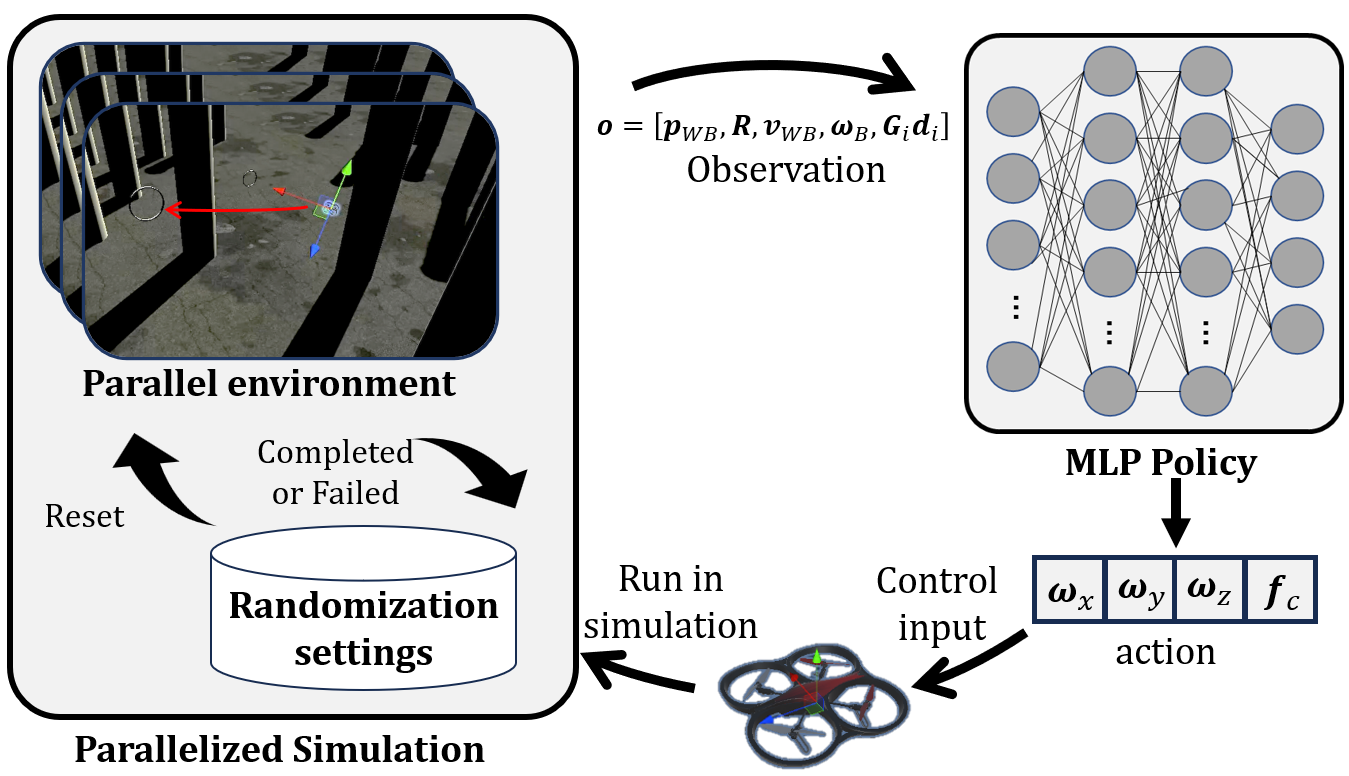}
        \caption{The proposed SRL framework}
        \label{farmework}
    %    \vspace{-0.5cm} 
\end{figure}

In the context of racing in cluttered environments, planning a collision-free trajectory with minimum-time traversal across a series of waypoints poses a formidable challenge. 
Achieving the objective of minimum time necessitates pushing the quadrotor to its physical extremes in terms of speed and acceleration. 
Furthermore, the existence of obstacles in three-dimensional space introduces highly non-convex optimization problems, making conventional methodologies impractical.  
The traditional framework often segregates planning and control, thereby increasing the risk of crashes due to inconsistencies in models.
This poses a substantial challenge to the agility of planners and the resilience of controllers.

Previous works either do not utilize the complete dynamics of the quadrotor \cite{liu2018search}, resulting in suboptimal and even dynamically infeasible solutions, 
or represent trajectories using polynomials or splines, which limit the positivity of control inputs \cite{Polynomial3}, reducing the aggressiveness of the trajectories.
While optimization-based techniques might provide optimal solutions \cite{Pham2018a_new_approach_to_time_optimal_path_parameterization}, they typically demand significant computational resources and time, making them less suitable for agile flight, particularly in environments rife with obstacles where computational demands are further compounded by non-convexity. 
Additionally, search-based methods \cite{liu2017search} encounter model mismatch problems due to the separation of planning and control.

Given the successful applications of deep reinforcement learning (DRL), we develop a DRL-based approach for autonomous drone racing in cluttered environments. 
DRL stands out as a promising alternative to traditional controller designs, harnessing the trial-and-error process to automatically optimize control strategies.
Its capacity to accommodate nonlinear dynamics and non-convex optimization objectives renders it well-suited for our application. 
Previous studies in autonomous drone racing have shown that neural network controllers trained with 
reinforcement learning (RL) outperform optimal control methods \cite{songscience}.

This work contributes a safe reinforcement learning (SRL) based method for racing quadrotors in cluttered environments.
In pursuit of minimizing flight time and ensuring obstacle avoidance in cluttered environments, a safety reward is proposed to encourage the quadrotor to maintain a safe distance from obstacles. 
Further, we introduce a terminal reward, activated upon task completion, to save flight time even further.
The proposed safety and terminal rewards are designed to facilitate performance that either matches or surpasses the capabilities of the state-of-the-art method presented in \cite{songscience} for achieving minimum-time flight within cluttered environments.
Extensive experimental analyses are undertaken, encompassing comparisons of flight time performance, evaluation of generalization capabilities to navigate in unseen environments, and ablation studies aimed at validating individual items of the proposed design.
Our experimental findings demonstrate that
\begin{itemize}
    \item Our method significantly improves the policy's obstacle avoidance capability in cluttered environments by introducing a safety reward enforcing obstacle avoidance. %at the price of flight time.
    \item By assigning terminal rewards (e.g., the objective of minimizing the total time) upon task completion, the negative impact by the safety reward can be mitigated. Experimental validation illustrates that this approach can achieve near-optimal agile and collision-free flights in highly complex environments.
    \item The SRL policy, trained with randomized environments of varying complexity levels, demonstrates robust generalization performance, achieving a commendable success rate of $66.7\%$ even in unseen, highly complex environments with dense obstacle configurations.
\end{itemize}

\section{RELATED WORK}

In traditional navigation results, methods for obstacle avoidance often decouple trajectory planning and control.
Given the minimum-time and collision-free trajectory, accurate trajectory tracking by the controller is instrumental for the quadrotor to navigate through the environment safely. 
These methods can be classified into polynomial and spline trajectory, search-based methods, and optimization-based methods.
The ultimate performance heavily relies on both the quality of the planned trajectory and the resilience of the controller.
Among these methods, the most popular approaches involve representing trajectories using polynomial \cite{fast-racing} 
and B-spline \cite{penin2018vision}. %,B2
The polynomial and spline methods simplify the states that need to be planned for by leveraging the differential flatness property \cite{ryu2011differential_flatness_based_robust_control}, and control inputs are obtained by taking their higher-order derivatives.
Yet, these methods generate smooth trajectories, which are suboptimal for minimum-time flight \cite{Polynomial3}.

Search-based methods translate trajectory planning into graph search problems, with the goal of optimizing discretized time intervals.
However, these methods \cite{liu2017search, liu2018search} face the curse of dimensionality, especially in high-dimensional spaces.
Furthermore, they rely on point-mass models for trajectory planning, which can compromise trajectory quality.
These methods have primarily been applied to path planning between two points, which contrasts with our multi-waypoint obstacle avoidance problem.

Optimization-based methods treat the time-optimal trajectory as a constrained optimization problem and solved it through nonlinear programming. 
These methods \cite{Pham2018a_new_approach_to_time_optimal_path_parameterization, fzyLearningHybrid,qin2023time} provide the optimal state and input control sequence at each step while satisfying dynamic constraints.
However, this approach often faces challenges in time allocation.
The work by \cite{CPC} introduces a complementary progress constraints (CPC) approach to address this issue, while also considering true actuator saturation.
Additionally, the work proposes a novel approach to computing time-optimal trajectories \cite{zzy} by fixing nodes with waypoint constraints and adopting separate sampling intervals for trajectories between waypoints.
This approach significantly reduces planning time compared to CPC but is not suitable for online scenarios with long trajectories.
Moreover, it does not consider environments with obstacles.

In recent years, learning-based methods have garnered extensive attention and achieved significant advancements in agile flight.
These methods \cite{cabrera2019gate, Cambridge2017real_single_image_flight} offer promising solutions to various challenges by training neural network policies to directly map high-dimensional observations to control commands.
For instance, researchers have utilized imitation learning to train neural network policies for agile flight \cite{highspeedinwild, IL2}, achieving high-speed trajectory tracking using learned policies,
and integrating topological path planning methods with DRL for obstacle avoidance.
Previous work \cite{songscience} has highlighted the superior performance of neural network controllers trained with RL over optimal control methods. 
This superiority stems from learning-based methods directly optimizing policies at the task level.
Nonetheless, existing work primarily focuses on obstacle-free environments and does not address scenarios with obstacles.
The work \cite{uzh_cluttered} combines DRL with topological path planning methods, leveraging the advantages of deep learning to overcome model mismatch issues.
In contrast, our approach does not require explicit trajectory representation but directly outputs control commands from observation states end-to-end.

\section{METHODOLOGY}

\subsection{Quadrotor Dynamics}

We model the quadrotor, actuated by four motors, as a $6$ degree-of-freedom rigid body of mass $\mathit{m}$ with a (diagonal) inertial matrix $\bm{\mathit{J}}$.
We refer to the quadrotor dynamics model presented in \cite{flightmare}, given as follows
\begin{equation}\label{eq:dynamic1}
        \begin{aligned}
                \bm{\dot p}_{WB} &= \bm{v}_{WB} \\
                \bm{\dot v}_{WB} &= \bm{q}_{WB} \odot \bm{c} - \bm{g} - \bm{RDR}^{T}\bm{v} \\
                \bm{\dot q}_{WB} &= \frac{1}{2}\bm{\Lambda}(\bm{\omega}_B) \cdot \bm{\dot q}_{WB} \\
                \bm{\dot \omega}_{B} &= \bm{J}^{-1}(\bm{\eta} - \bm{\omega}_{B} \times \bm{J\omega}_{B}) 
        \end{aligned}
\end{equation}
where $\odot$ denotes the quaternion multiplication,
$\bm{ p}_{WB}$ and $\bm{v}_{WB}$ represent the position and linear velocity of the quadrotor in the reference frame $WB$. 
The quaternion $\bm{q}_{WB}$ represent for the attitude of the quadrotor, $\bm{\omega}_B$ is the body rates in the body frame $B$, $\bm{R}$ is the rotation matrix, and $\bm{D} = {\rm diag}(d_{x}, d_{y}, d_{z})$ is a constant diagonal matrix collecting the rotor drag coefficients, which is a linear effect in velocity.
Moreover, $\bm{\Lambda}(\bm{\omega}_B)$ is a skew-symmetric matrix, $\bm{c}=[0,0,c]$ is the mass-normalized thrust vector and $\bm{g}=[0,0,-g_{z}]^{T}$ is the gravity vector with $g_{z}=9.81\,{\rm m/s}^2$. The conversion of the four rotor thrusts $[f_1, f_2, f_3, f_4]$ to the mass-normalized thrust $c$ and the body torques $\bm \eta$ are given by
\begin{equation}\label{eq:dynamic2}
        \begin{aligned}
                \bm{\eta} &= 
                \begin{bmatrix}
                        \frac{l}{\sqrt{2}}(f_1 - f_2 - f_3 + f_4) \\
                        \frac{l}{\sqrt{2}}(-f_1 - f_2 + f_3 + f_4) \\
                        \kappa(f_1 - f_2 + f_3 - f_4) \\
                \end{bmatrix} \\
                c &= \frac{1}{m}(f_1 + f_2 + f_3 + f_4) \\         
        \end{aligned}
\end{equation}
where $l$ is the arm length of the quadrotor, $m$ is the its mass, and $\kappa$ is the torque constant.
The thrust dynamics of each individual rotor is approximated as a first-order system $\dot f_i = (f_{d,i} - f_i)/k_m$, where $f_{d,i}$ is the desired thrust of rotor $i$ and $k_m$ is the time-delay constant. We employ
a $4$th-order Runge-Kutta method for integrating the dynamic equations.

\subsection{Reinforcement Learning for Minimum-time Flight without Collisions}\label{sec: Probelm Formulation}

We formulate the autonomous drone racing problem in obstacle-rich environments within the RL framework. 
In this context, we model the drone racing task as a Markov decision process (MDP) defined by the $6$-tuple $(\mathcal{S}, \mathcal{A}, \mathcal{P}, r, \rho_0, \gamma)$ \cite{sutton1999reinforcement},
where $\mathcal{S}$ represents the state space, $\mathcal{A}$ is the action space,
$\mathcal{P}$ is the state transition function which defines the probability of transition from state $s\in\mathcal{S}$ to state $s'\in\mathcal{S}$ by taking action $a\in\mathcal{A}$, the reward function $r$ determines that the quadrotor will receive an immediate reward $r(s,a)$ upon executing action $a$ in state $s$.
The quadrotor starts from a state $s_0 \in \mathcal{S}$ drawn from the initial state distribution $\rho_0$. 
At each step $k$, an action $a(k) $ is randomly sampled from a policy $\pi_{\theta}: \mathcal{S}\to \mathcal{A}$ parameterized by a deep neural network (DNN) with weight parameters collected in $\theta$. 

Our goal is to find the optimal policy $\pi_\theta^*$, namely, the weight parameters $\theta$ of the DNN policy $\pi_{\theta}$, by maximizing the expected accumulative discounted reward as follows
\begin{equation}\label{eq:rl}
        \pi_{\theta}^{*} = \arg\max_{\pi_{\theta}}\mathbb{E}\Bigg[ \sum_{k=0}^{\infty}\gamma^kr(k) \Bigg].
\end{equation}

\subsubsection{Reward function}\label{sec: reward function}
Training RL policies directly with the objectives of minimizing total flight time and penalizing collisions appears intuitive. 
However, total flight time and collision penalties represent sparse rewards, which make the learning difficult and often leads to suboptimal solutions. 
Furthermore, as the environment complexity escalates, training becomes increasingly challenging.
A feasible strategy to address this challenge involves employing a proxy reward that closely approximates the true performance objective while offering feedback to the quadrotor at each step.

Rewards computed based on the path progress serve as an alternative to the minimum-time flight reward.
The gate progress reward, as utilized in \cite{songscience}, provides a straightforward and efficient optimization objective to replace minimum-time flight while circumventing sparse rewards.  
Our approach extends the gate progress reward and introduces additional ones to enhance the efficacy of minimum-time flight in cluttered environments.
As an ingredient of our reward, the gate progress reward $r_p(k)$ facilitates the quadrotor to find the shortest time path, given by
\begin{equation}\label{eq:gaetprogress}
        r_p(k) = \lambda_{1}(||g(k) - p(k-1)|| - ||g(k) - p(k)||) - \lambda_{2}||\omega(k)||
\end{equation}
where $g(k)$ and $p(k)$ represent the target point (e.g. the target gate center) and the position of the quadrotor at step $k$, respectively.
The term $||g(k) - p(k-1)|| - ||g(k) - p(k)||$ guides the quadrotor towards the center of the current target gate;
once it reaches the gate, its next gate in racing is set as the target point.
Moreover, $\lambda_{2}||\omega (k)||$ denotes a penalty on the body rate, where $\lambda_{2}$ is a small coefficient.

Additionally, we introduce a safety reward to prevent collisions in cluttered environments, by progressively keeping the quadrotor away from obstacles within an observation range using exponential functions, defined as follows
\begin{equation}\label{eq:safety reward}
        r_s(k) = \lambda_{3}\sum_{i}^{N}e^{-||d_{i}(k)||}
\end{equation}
where $\lambda_3$ is a coefficient that adjusts the aggressiveness of the trajectory, $N$ is designated as the nearest $N$ obstacles within the quadrotor's observation range, 
and $||d_{i}(k)||$ denotes the distance between the quadrotor and the $i$-th obstacle.

Therefore, the overall reward at each step $k$ is given by
\begin{equation}\label{eq:terminal reward}
        r(k) = r_p(k) + r_s(k) + \begin{cases}
                r_c, &  {\rm if~collision} \\
                \lambda_4 T, &{\rm if~completed} \\
                0, & {\rm else} \\
        \end{cases}
\end{equation}
where $r_c$ and $\lambda_4 T$ are the terminal rewards depending on whether collision occurs or the racing task is successfully completed. 
In our experiments, a substantial negative reward of $r_c=-30$ was applied in the event of a collision. 
Task completion is deemed unsuccessful if the quadrotor either collides or fails to reach the designated target point. The variable $T$ signifies the total time required by the quadrotor to complete the task. Additionally, the reward $\lambda_4 T$ is introduced to penalize trajectories with excessively prolonged completion times

\subsubsection{Policy architecture}
The policy network consists of two layers, which take observations $\bm{o}$ as input and directly map them to the quadrotor's action $\bm{a}$. 
The observation space of the DNN policy comprises three components: the state of the quadrotor, the positions of the next few target points, and the distances between the quadrotor and $N$ obstacles.
Following  \cite{hwangbo2017control}, we utilize rotation matrices to represent the attitude of the quadrotor, thereby avoiding ambiguities and discontinuities in gesture expression.
The quadrotor's observation is defined as $\bm{o}(k) = [\bm{p}_{WB}(k), \bm{R}(k), \bm{v}_{WB}(k), \bm{\omega}_B(k), \bm{G}_{j}(k), ||d_{i}(k)||]$, 
where $\bm{p}_{WB}(k), \bm{R}(k), \bm{v}_{WB}(k)$, and $\bm{\omega}_B(k)$ are the quadrotor's position, rotation matrix, linear velocity, and body rate, 
and $\bm{G}_{j}(k)$ collects the positions of the next $j$ target points with the current target point included (we set $j = 2$ in our experiment). 
The hyper-parameter $j$ can take any value equal to or less than the total number of target points.
Additionally, $||d_{i}(k)||$, where $i \in [1,\ldots,N]$, represents the Euclidean distance between the quadrotor and the $i$-th obstacle.

The action is defined as $\bm{a}(k)=[f^{\rm total}(k), \bm{\omega}(k)]$, which consists of the desired collective thrust $f^{\rm total}(k)$ and the body rate $\bm{\omega}(k)$. 
Subsequently, a low-level controller is utilized to execute the action commands.
Notably, this approach has exhibited robust sim-to-real transferability in learning-based control policies \cite{Kaufmann2022a_benchmark_comparison}.

\subsection{Policy Training}
Improving the training efficiency and generalization of DRL method remains a formidable challenge in robotics research.
Previous studies have suggested solutions, including parallel training, segmented training \cite{uzh_cluttered}, and domain randomization \cite{highspeedinwild}.
To bolster the generalization performance, we introduce a set of modifications to the training environment, as elaborated below.

\subsubsection{Parallel environment simulation}
To train the policy in simulation, we utilize the open-source Flightmare simulator \cite{flightmare}, which can simulate several hundred quadrotors with a flexible physics engine.
Specifically, $100$ environments are designed for training RL policies. 
Each environment allows for different initialization settings and variations, significantly enhancing data diversity and saving time for data collection.

\subsubsection{Representation of collision}
Creating environments in the physics engine based on Flightmare involves establishing interaction rules between the environment and the drone, such as collision detection. 
In our training, obstacles, such as spherical, cylindrical, and cubical obstacles, are simplified as single or multiple spherical obstacles with a radius of $r$. 
A safety distance $d_{\text{safe}} = r + l + \epsilon_{\text{safe}}$ is defined to assess potential collisions, with $\epsilon_{\text{safe}}$ denoting a safety threshold.
If the distance between the quadrotor and any obstacle falls below the safety distance, i.e., $d_{i,k} < d_{\text{safe}}$, or if the quadrotor surpasses the world boundaries, it is signaled as a collision, prompting a reset of the quadrotor's position.

\subsubsection{Domain randomization}
Domain randomization has emerged as an effective strategy for mitigating environmental and model uncertainties.
To enhance the generalization performance and robustness of the RL planner, we incorporate domain randomization into our methodology.
Throughout the training phase, the initial states of the quadrotor, the positions of waypoints, and the complexity of the environment (i.e., clutter) are randomly generated.
The quadrotor parameters, such as linear drag coefficient and thrust mapping coefficient, are randomly sampled around calibrated values at each reset to ensure the policy's robustness against unknown and potentially stochastic aerodynamic effects.

\section{EXPERIMENTS}

In this section, a lot of tests were conducted to compare the proposed SRL policy with state-of-the-art methods and to evaluate its robustness and generalization capabilities in uncertain and unseen experimental scenarios.
Additionally, a series of ablation studies were performed to validate the design choices of the proposed method.

\begin{figure*}[thpb]
        \centering
        %\framebox{\parbox{3in}{We suggest that you use a text box to insert a graphic (which is ideally a 300 dpi TIFF or EPS file, with all fonts embedded) because, in an document, this method is somewhat more stable than directly inserting a picture.}}
        \includegraphics[scale=0.5]{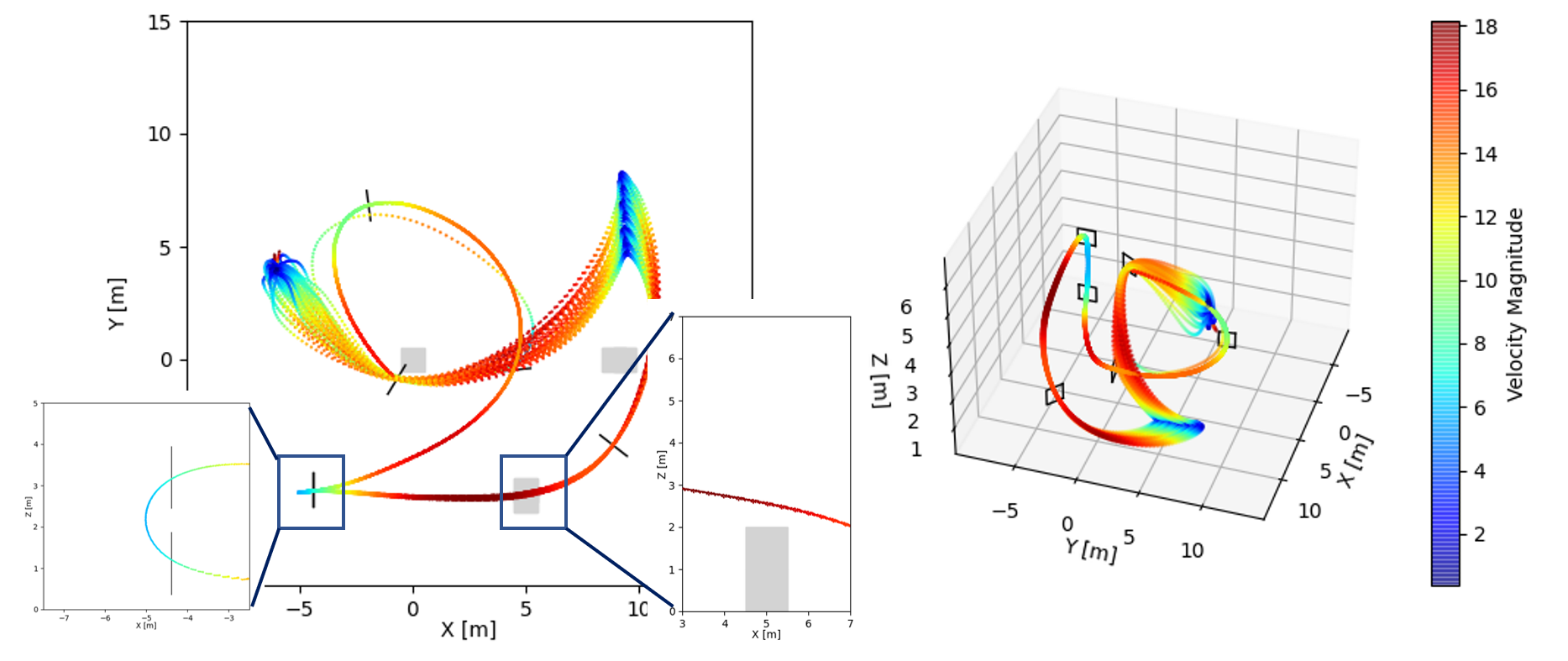}
        \caption{
                The trajectories generated by our method in Split-S with obstacles are depicted in the figure. The light gray area represents obstacles. During the flight, the quadrotor needs to pass through $7$ waypoints, with one of them being randomly initialized in position for each flight.
                }
        \label{pic:split-s}
        \vspace{0cm} 
\end{figure*}

\begin{figure*}[thpb]
        \centering
        %\framebox{\parbox{3in}{We suggest that you use a text box to insert a graphic (which is ideally a 300 dpi TIFF or EPS file, with all fonts embedded) because, in an document, this method is somewhat more stable than directly inserting a picture.}}
        \includegraphics[scale=0.5]{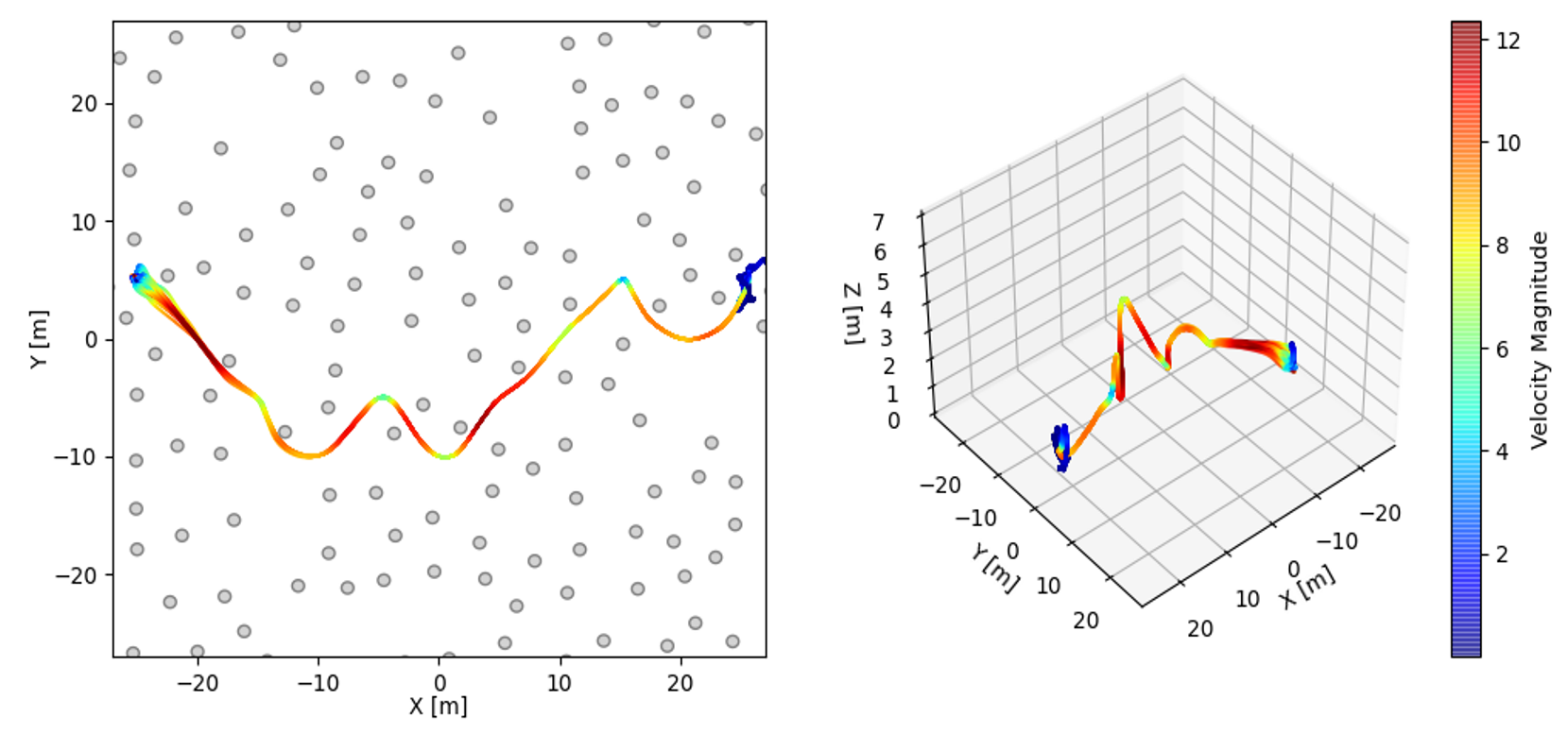}
        \caption{
                The trajectories generated by our approach in a level $1$ forest environment navigate through nine waypoints while effectively avoiding obstacles. The light gray area in the figure represents the obstacles.
                }
        \label{pic:forest}
        \vspace{0cm} 
\end{figure*}

\subsection{Experiment Setup}\label{sec: Experimental Setup}
The proximal policy optimization (PPO) algorithm \cite{PPO} stands out for its robust performance and easy tuning process, making it a favorable choice among RL researchers.
In our study, we leveraged the PPO algorithm to train the SRL policy that generates the quadrotor's action.
PPO relied on Stable-Baseline$3$ \cite{SB3}, a trusted tool for RL implementations.
Several cluttered environments were designed based on Flightmare, allowing for the concurrent simulation of hundreds of drones and environments.
This approach significantly bolstered the efficiency of data collection.

The scenarios, Split-S with obstacles and Forest with varying density levels, were designed to evaluate the performance of our approach.
In both scenarios, the mission requires the quadrotor to traverse a sequence of designated waypoints from the starting point to the terminal while avoiding the obstacles.
The distribution of waypoints on the Split-S track was generated similar to \cite{songscience}, while the positions of waypoints in the Forest environment were given in Table \ref{Position of waypoints}.
In contrast to the Split-S track, the Forest environment presented a longer track with more obstacles. 
Two experiments were conducted to respectively evaluate the performance of flight time and the generalization capabilities of the policy using these environments, along with several ablation studies to validate our design.
At the beginning of each training, the state of the quadrotor or the elements in the environment were randomly initialized.
And all tests collected 1000 trajectories for evaluation.
Two metrics, average time and crash ratio, were used to evaluate the policy's performance.
For Split-S, average time represents the mean lap time, while for Forest, it represented the mean time required to reach the endpoint.
Unless noted otherwise, all experiments were conducted using a quadrotor configuration with diagonal inertia matrix $\bm{J} = [0.0025, 0.0021, 0.0043]$, rotor-drag coefficients $d_x = 0.26$, $d_y = 0.28$, $d_z = 0.42$, arm length $l = 0.15$ and torque constant $\kappa = 0.022$.
Additionally, the parameters in the reward function were set as follows: $\lambda_1 = 1$, $\lambda_2 = 0.01$, $\lambda_3 = -0.05$, and $\lambda_4 = -1$.

\begin{table}[t]
        \setlength{\abovecaptionskip}{0cm} 
	\small
        \caption{Positions of waypoints.}
        \label{Position of waypoints}
        \begin{center}
                \begin{tabular}{c c}
                        \toprule
                        Waypoint number & Position \\
                        \midrule
                        $1$ & [$-20.0$, $0.0$, $1.12$] \\
                        $2$ & [$-15.0$, $-5.0$, $1.12$] \\
                        $3$ & [$-10.0$, $-10.0$, $1.12$] \\
                        $4$ & [$-5.0$, $-5.0$, $1.12$] \\
                        $5$ & [$0.0$, $-10.0$, $3.53$] \\
                        $6$ & [$5.0$, $-5.0$, $1.12$] \\
                        $7$ & [$10.0$, $0.0$, $3.53$] \\
                        $8$ & [$15.0$, $5.0$, $3.53$] \\
                        $9$ & [$20.0$, $0.0$, $1.12$] \\
                        \bottomrule
                \end{tabular}
        \end{center}
        \vspace{-0.0cm} 
\end{table}

\subsection{Minimum-time Flight}\label{sec: Minimum-time Flight}

To benchmark the performance of our method in capability of flight time, a comparative analysis between our approach and the recently proposed method \cite{songscience} was conducted.
Since the RL method in \cite{songscience} is not open-sourced, we reproduce the algorithm based on the information provided in \cite{songscience} and achieve consistent results  with those reported.
In this experiment, we only randomized the initial state of the drone and one of waypoints at the beginning of each flight, including its position, linear velocity, and orientation.
Specifically, the starting point of the quadrotor $[p_{WB,x}, p_{WB,y}, p_{WB,z}]$ was determined, and it wsa varied within the range $[p_{WB,x}\pm\Delta x, p_{WB,y}\pm\Delta y, p_{WB,z}\pm\Delta z]$, where $\Delta x$, $\Delta y$, and $\Delta z$ are randomly initialized between $[-1, 1]$. 
Similarly, the linear velocity and orientation were also randomly initialized between $[-1, 1]$.
It is worth noting that for the Forest environment, we did not randomize the obstacle density during training, and the environments used for testing were the same as those during training.

The results, as shown in Table \ref{minimum-time}, indicate promising performance for our method.
Additionally, Figure \ref{pic:split-s} illustrates the $100\%$ success rate achieved for both obstacle avoidance and the completion of tasks involving multiple waypoints in Split-S with obstacles.
Our method demonstrates performance that either exceeds or closely approaches that of state-of-the-art methods in both environments.
The Forest environment presents greater challenges due to its increased number of obstacles and longer track length.
Despite these complexities, our method achieves performance closely aligned with state-of-the-art benchmark, while ensuring collision-free navigation throughout the entire trajectory.
The improvement in performance can be attributed to the terminal reward $\lambda_4 T$ incorporated into our reward function, as demonstrated in Section \ref{sec: ablation studies}.

\begin{table}[t] 
    \setlength{\abovecaptionskip}{0cm} 
	\setlength{\belowcaptionskip}{0cm}
	\small
        \caption{Comparison of baseline and our RL method.}
        \label{minimum-time}
        \begin{center}
            \renewcommand{\arraystretch}{1.3}
            \begin{tabular}{c c c c}
                \toprule
                Method & Environment &Time (s) &  Crash ratio ($\%$) \\
                \midrule
                \multirow{2}*{\cite{songscience}} & Split-S with obstacles & $7.11$ & $0$\\ 
                                                  & Forest &$7.34$ & $90.20$\\ \cline{1-4}
                \multirow{2}*{Ours} & Split-S with obstacles & $\mathbf{6.49}$  & $0$\\ 
                                    & Forest & $ \mathbf{7.31}$ & $0$\\
                \bottomrule
            \end{tabular}
        \end{center}
        \vspace{-0.0cm} 
\end{table}

\subsection{Performance in Unseen Environments}

To evaluate the robustness and generalization of the proposed SRL policy, in addition to randomly initializing the quadrotor's state as described in \ref{sec: Minimum-time Flight}, we also changed the complexity of the environment.
Specifically, $100$ environments with varying complexity, representing different densities of obstacles, were created to train the policy. 
During testing, we randomly generated three unseen environments and categorized them into three levels based on the density of obstacles: Level $1$ (obstacle spacing approximately $5m$), Level $2$ (obstacle spacing approximately $3-5m$), and Level $3$ (obstacle spacing approximately $1-3m$).

The outcomes are depicted in Table \ref{forest test}. 
For each level, we executed tests on $1000$ trajectories..  
Figure \ref{pic:forest} showcases the performance of our method in Level $1$.
The trained neural network achieves a success rate of $66.7\%$ in the most challenging environment, which is not seen during training. 
The reduction in average time can be attributed to the heightened complexity of the environment.

\begin{table}[h]
        \setlength{\abovecaptionskip}{0cm} 
	\setlength{\belowcaptionskip}{0cm}
	\small
        \caption{Evaluation results in unseen environments.}
        \label{forest test}
        \begin{center}
                \begin{tabular}{c c c}
                        \toprule
                        Forest environment & {\parbox{2.5cm}{\centering Average time (s)}} & Crash ratio (\%)\\
                        \midrule
                        Level $1$ \quad & $7.31$ & $0$ \\
                        Level $2$ \quad & $7.33$ & $0$ \\
                        Level $3$ \quad & $7.46$ & $33.33$\\
                        \bottomrule
                \end{tabular}
        \end{center}
        \vspace{-0.0cm}
\end{table}

\subsection{Ablation Studies}\label{sec: ablation studies}

We conducted ablation experiments to validate our method, focusing primarily on the design of our reward function in both Forest and Split-S with obstacles.
In these experiments, there were no changes made to the elements in the environment during training and testing, and the environments used for testing were the same as those during training.
Only the initial state of the drone was randomized during both training and testing phases.

\subsubsection{Safety reward}

A safety reward, mentioned in Section \ref{sec: Probelm Formulation}, was designed to encourage the drone to avoid obstacles.
A smaller value of $\lambda_3<0$ corresponded to more assertive trajectories.
This experiment aimed to validate the impact of the safety reward on the policy.
The result as shown in Table \ref{ablation Studies for safety reward}.
In the Forest environment, "Average Time" signifies the completion time, whereas in the Split-S environment, it represents the lap time.
The Forest environment presents a higher degree of complexity compared to Split-S.  
Consequently, in the absence of the safety reward, the trained policy carries a risk of collisions. 
This underscores the beneficial influence of the safety reward introduced by our approach on obstacle avoidance in intricate environments.
\begin{table}[t]
    \setlength{\abovecaptionskip}{0cm} 
	\setlength{\belowcaptionskip}{-0.2cm}
	\small
    \caption{Ablation studies testing the efficacy of the safety reward.}
    \label{ablation Studies for safety reward}
    \begin{center}
        \renewcommand{\arraystretch}{1.3}
        \begin{tabular}{c c c c}
            \toprule
            Environment & Safety reward & Crash ratio ($\%$) \\
            \midrule
            \multirow{2}*{Forest} & \textbf{\checkmark} & $0$ \\ 
                                    & \ding{55} & $1.96\%$ \\ \cline{1-3}
            \multirow{2}*{Split-S with obstacles} & \textbf{\checkmark} & $0$ \\ 
                                                    & \ding{55} & $0$ \\
            \bottomrule
        \end{tabular}
    \end{center}
    \vspace{-0.0cm} 
\end{table}

\subsubsection{Terminal reward}
In the early stages of training, the quadrotor triggered the terminal reward $r_c$ in \eqref{eq:terminal reward} due to flying out of the world boundaries or colliding with obstacles because of its randomly initialized policy.
This reward was assigned a substantial negative value to prevent the quadrotor from revisiting similar actions.
In this experiment, our focus was on studying the impact of the reward $\lambda_4 T$ on the SRL policy.
The results of the experiment are presented in Table \ref{ablation Studies for terminal reward}.
When eliminating the term $\lambda_4 T$, the policy exhibits subpar performance in terms of task completion time. 
The experimental results demonstrate that our design has a positive impact on minimum-time flight.
\begin{table}[h]
        \setlength{\abovecaptionskip}{0cm} 
	\setlength{\belowcaptionskip}{-0.0cm}
	\small
        \caption{Ablation studies for testing the efficacy of the terminal reward.}
        \label{ablation Studies for terminal reward}
        \begin{center}
            \renewcommand{\arraystretch}{1.3}
            \begin{tabular}{c c c}
                \toprule
                Environment & $\lambda_4 T$ & Average time (s)\\
                \midrule
                \multirow{2}*{Forest} & \textbf{\checkmark} & \textbf{$7.31$} \\
                                    & \ding{55} & $8.57$ \\ \cline{1-3}
                \multirow{2}*{Split-S with obstacles} & \textbf{\checkmark} & \textbf{$6.49$} \\
                                                    & \ding{55} & $7.12$ \\
                \bottomrule
            \end{tabular}
        \end{center}
        \vspace{-0.0cm} 
\end{table}

\section{Conclusions}

In this paper, we presented a novel learning-based methodology aimed at training an SRL policy to guide a quadrotor through cluttered environments, ensuring minimal flight time and collision-free navigation. To this end, novel safety and terminal rewards as well as training procedures are designed. 
Our experimental results showcase the robustness and generalization capabilities of the proposed approach.
Additionally, ablation analyses underscore the positive contributions of each component within our framework. 
These findings indicate that SRL shows significant potential for high-speed flight and obstacle avoidance.

\addtolength{\textheight}{-6cm}   % This command serves to balance the column lengths
                                  % on the last page of the document manually. It shortens
                                  % the textheight of the last page by a suitable amount.
                                  % This command does not take effect until the next page
                                  % so it should come on the page before the last. Make
                                  % sure that you do not shorten the textheight too much.

%%%%%%%%%%%%%%%%%%%%%%%%%%%%%%%%%%%%%%%%%%%%%%%%%%%%%%%%%%%%%%%%%%%%%%%%%%%%%%%%

%%%%%%%%%%%%%%%%%%%%%%%%%%%%%%%%%%%%%%%%%%%%%%%%%%%%%%%%%%%%%%%%%%%%%%%%%%%%%%%%

%%%%%%%%%%%%%%%%%%%%%%%%%%%%%%%%%%%%%%%%%%%%%%%%%%%%%%%%%%%%%%%%%%%%%%%%%%%%%%%%

%%%%%%%%%%%%%%%%%%%%%%%%%%%%%%%%%%%%%%%%%%%%%%%%%%%%%%%%%%%%%%%%%%%%%%%%%%%%%%%%

\bibliographystyle{IEEEtran}
\bibliography{./bibtex/paper}

\end{document}